\documentclass[runningheads]{llncs}

\usepackage{eccv}
\usepackage{pdfpages}

\usepackage{eccvabbrv}

\usepackage{graphicx}
\usepackage{booktabs}

\usepackage[accsupp]{axessibility}  
\usepackage{booktabs}
\usepackage{multirow}
\usepackage{graphicx}      
\usepackage[table]{xcolor} 
\usepackage{amssymb}       
\definecolor{LightGray}{gray}{0.93}
\usepackage{wrapfig}
\usepackage{subcaption}
\usepackage{booktabs}
\usepackage{multirow}
\usepackage{siunitx}
\usepackage{hyperref}

\usepackage{orcidlink}

\begin{document}

\title{LatentPilot: Scene-Aware Vision-and-Language Navigation by Dreaming Ahead with Latent Visual Reasoning} 


\author{
Haihong Hao\inst{1}\textsuperscript{*} \and
Lei Chen\inst{1}\textsuperscript{*} \and
Mingfei Han\inst{2}\textsuperscript{*} \and
Changlin Li\inst{3} \and
Dong An\inst{4} \and
Yuqiang Yang\inst{5} \and
Zhihui Li\inst{1} \and
Xiaojun Chang\inst{1}
}

\institute{
University of Science and Technology of China \and
MBZUAI \and
Stanford University \and
Amap, Alibaba Group \and
Shanghai AI Laboratory
\textsuperscript{*}Equal contribution.
}

\maketitle
\vspace{-15pt}
\begin{center}
\footnotesize{
Project Page: \url{https://abdd.top/latentpilot/}}
\end{center}
\vspace{10pt}

\begin{abstract}
 Existing vision-and-language navigation (VLN) models primarily reason over past and current visual observations, while largely ignoring the future visual dynamics induced by actions. As a result, they often lack an effective understanding of the causal relationship between actions and how the visual world changes, limiting robust decision-making. Humans, in contrast, can “imagine” the near future by leveraging action–dynamics causality, which improves both environmental understanding and navigation choices. Inspired by this capability, we propose LatentPilot, a new paradigm that exploits future observations during training as a valuable data source to learn action-conditioned visual dynamics, while requiring no access to future frames at inference. Concretely, we propose a flywheel-style training mechanism that iteratively collects on-policy trajectories and retrains the model to better match the agent’s behavior distribution, with an expert takeover triggered when the agent deviates excessively. LatentPilot further learns visual latent tokens without explicit supervision; these latent tokens attend globally in a continuous latent space and are carried across steps, serving as both the current output and the next input, which enabling the agent to “dream ahead” and reason about how actions will affect subsequent observations. Experiments on R2R-CE, RxR-CE, and R2R-PE benchmarks achieve new SOTA results, and real-robot tests across diverse environments demonstrate LatentPilot’s superior understanding of environment–action dynamics in scene.
\end{abstract}



\section{Introduction}
Vision-and-Language Navigation (VLN) requires an embodied agent to move through a 3D environment by following natural-language instructions and reaching a target location~\cite{anderson2018vln,ku2020rxr,krantz2020beyond}. Solving VLN demands more than aligning language with the current visual observation. The agent must accumulate spatial memory, interpret scene semantics, and correct itself under uncertainty during sequential decision making. Crucially, VLN is not a static “perceive then act” problem. It is an interactive process in which every action changes what will be observed next. Therefore, robust navigation is tightly linked to the ability to understand the environment and to anticipate how it will evolve under the agent’s actions, namely an ability to imagine near-future observations.

\begin{figure}[t!] 
    \centering
    \includegraphics[width=0.94\linewidth]{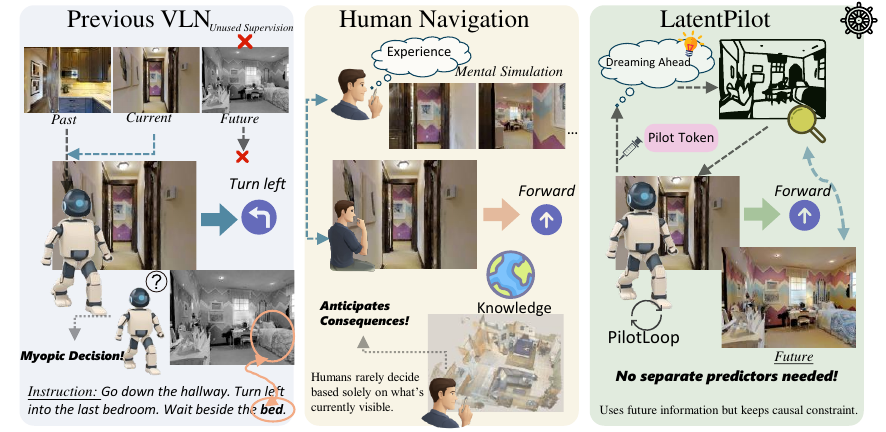}
    \vspace{-2mm}
    \caption{
Prior VLN pipelines optimize next-action prediction from past/current observations, leaving future observations in trajectories largely unused as supervision. LatentPilot uses future observations only as training-time privileged supervision to internalize action-conditioned visual dynamics while inference remains strictly causal and requires neither future frames nor an external world-model rollout.
    }
    \vspace{-3.5mm}
    \label{fig:observation}
\end{figure}

Most existing VLN approaches formulate navigation as conditional policy learning~\cite{anderson2018vln,krantz2020beyond}. Given an instruction and past or current observations, the model predicts the next action, typically trained via imitation learning on expert trajectories~\cite{ross2011dagger} that fit “what to do now.” As shown in Fig.~\ref{fig:observation}, although training trajectories inherently contain rich future frames that reflect the consequences of actions, mainstream training pipelines usually treat them only as later observations in a sequence. They are seldom used as explicit supervision to shape representations that are predictive and anticipatory at the current step.
A complementary line of work equips agents with explicit lookahead via world models or imagination modules that predict or synthesize future observations and then plan, rerank, or prompt decisions accordingly (e.g., Pathdreamer~\cite{koh2021pathdreamer}, future-view generation objectives~\cite{li2023futureview}, diffusion-based imagination for VLN~\cite{huang2025vista,huang2025vistav2}, and scene-imagination prompting pipelines~\cite{zhao2025imaginenav,perincherry2025vlnimagine}, as well as world-model-based VLN-CE frameworks~\cite{yao2025navmorph}). While effective, these approaches typically externalize imagination as separate predictors, planners, or sampling procedures, introducing additional computation and exposing the agent to compounding prediction errors and policy--model mismatch.
In contrast, we argue that it is beneficial to internalize imagination: distilling action--observation causality into the navigator itself so that anticipatory reasoning becomes an amortized, end-to-end capability of the same backbone that makes decisions.

 If decisions depend only on past and current observations, policies can become myopic, behaving in a “step and see” manner, and may commit to erroneous branches that are difficult to recover from in complex layouts. In contrast, if a model can anticipate how a candidate action will shape near-future observations, it becomes more likely to avoid inefficient exploration, reduce collisions and backtracking, and improve overall stability.Although future observations are not available at decision time, they are fully recorded in offline training trajectories. This raises a central question: can we use these recorded future observations during training to learn how actions change future visual inputs, while keeping inference strictly causal?

When moving in the real world, people rarely make decisions purely on what is visible at the moment. Instead, they perform quick mental simulation before acting, for example, whether turning left will reveal a corridor or a doorway, or whether moving forward will lead to an open area or a dead end. This ability is not innate. It is acquired through experience and reflection on the true consequences of past actions, gradually forming an internal forward model that supports pre-action simulation. By comparison, many VLN methods are effective at extracting cues from past and current observations, yet they rarely model the causal relation between actions, environment change explicitly. Fig. ~\ref{fig:observation} suggests a more natural direction. During training, trajectories already include what the agent actually sees after taking actions. If these future observations are used as supervision, the “hindsight” consequences can be converted into “foresight” representations, allowing the model to learn more anticipatory decision making without requiring access to any future images at test time.

Building on this insight, we propose \textbf{Latent} Visual Reasoning–based Dreaming ahead \textbf{P}olicy for Scene-Aware Vision-and-Language Navigation, \textbf{LatentPilot}. It is an end-to-end VLM navigator that maintains a step-propagated visual latent, termed the Pilot Token, to carry compact near-future imagination across decisions. Unlike explicit imagination/world-model pipelines that rely on separate predictors or rollouts at inference~\cite{huang2025vista,huang2025vistav2,zhao2025imaginenav,yao2025navmorph}, LatentPilot amortizes lookahead into the same decision backbone via the step-propagated Pilot Token.
The key observation is that while future observations are unavailable before an action is executed at test time, they are naturally recorded in pre-collected navigation trajectories. We therefore treat these future views as training-only privileged supervision, encouraging the model to produce Pilot Tokens that are predictive of the visual evidence it will encounter after acting. Importantly, LatentPilot remains strictly causal at inference: it never accesses future frames, and instead relies on the current observation, the instruction, and the propagated Pilot Token for decision making. To reduce the mismatch between offline training and interactive deployment, we further adopt a flywheel-style learning loop called PilotLoop that repeatedly collects rollouts and updates the model~\cite{ross2011dagger}. 

To summarize, our contributions are as follows: \begin{itemize} 
    \item LatentPilot framework: We introduce an end-to-end LMM-based navigator with a propagated visual latent (Pilot Token) that internalizes future-aware reasoning within the decision backbone, without requiring any external imagination module, world-model rollout, or planning procedure at inference.
    \item We introduce a training-only privileged supervision objective that leverages future observations in trajectories to learn action-conditioned visual dynamics in latent space, effectively converting trajectory hindsight into test-time foresight while keeping inference strictly causal. It does not introduce future information leakage at test time.  
    \item We demonstrate consistent gains in both simulator benchmarks like R2R-CE, RxR-CE, and R2R-PE, and real-robot settings, indicating strong generalization and practical potential.
\end{itemize}
\section{Related Work}
\textbf{Vision-and-Language Navigation} (VLN) studies embodied agents that navigate in 3D environments by following natural-language instructions and reaching a target location. Early benchmarks such as Room-to-Room (R2R) are built on the discretized viewpoints and navigation graphs of Matterport3D\cite{anderson2018vln,chang2017matterport3d}. RxR extends this setting with multilingual instructions and denser annotations, further increasing coverage and difficulty \cite{ku2020rxr}. VLN-CE moves evaluation to continuous environments, making the task more deployment-oriented \cite{krantz2020beyond}. Large-scale indoor 3D datasets such as HM3D substantially increase scene diversity for training and evaluation \cite{ramakrishnan2021hm3d}. ScaleVLN synthesizes large-scale instruction–trajectory pairs via data generation \cite{wang2023scalevln,xia2018gibson}. VLN-PE systematically characterizes performance degradation caused by physical and visual disparities and provides a more realistic evaluation platform across different robot embodiments \cite{wang2025vlnpe}.
 ETPNav performs online topological mapping and separates long-horizon planning from low-level control to improve executability in continuous environments \cite{an2023etpnav}. BEVBert advances multimodal pretraining from a map-centric perspective, strengthening spatially aware representations for navigation \cite{an2023bevbert}. NaVid formulates continuous navigation as video-conditioned next-step planning and outputs actions from streaming visual observations \cite{zhang2024navid}, while Uni-NaVid further unifies multiple embodied navigation tasks under a video-based VLA formulation \cite{zhang2025uninavid}.

Some works build world models that synthesize unobserved or future views to support planning or action evaluation, such as Pathdreamer \cite{Koh_2021_ICCV} and NavMorph \cite{Yao_2025_ICCV}. Other approaches treat future views as learning targets and introduce auxiliary objectives that predict or align with next-step observation semantics, such as future-view image semantics generation \cite{Li_2023_CVPR}. While these methods demonstrate the value of future-aware signals, anticipatory reasoning is often implemented as separate predictors, generative models, or inference-time rollouts and fusion procedures, which may introduce additional computation and policy--model mismatch during closed-loop execution. In contrast, our work aims to integrate this capability into a single end-to-end navigator.

\textbf{Latent reasoning} shifts multi-step computation into continuous latent space rather than explicit chain-of-thought text, reducing token cost while retaining iterative internal processing. In language models, Think Before introduces pause tokens for extra latent computation \cite{goyal2023think}, Quiet-STaR learns implicit rationales \cite{zelikman2024quietstar}, and Coconut feeds hidden states back as reasoning states \cite{hao2024coconut}. Related efforts further distill or compress latent reasoning, including CoDi \cite{shen2025codi}, SIM-CoT \cite{wei2025simcot}, and Think Silently, Think Fast \cite{tan2025thinksilently}. In multimodal settings, latent reasoning has expanded from language-only states to visual and joint vision--language latents. LVR reasons autoregressively in visual embedding space \cite{li2025lvr}; Reasoning in the Dark interleaves latent vision--text reasoning \cite{chen2025reasoningdark}; CoCoVa performs iterative multimodal refinement through continuous cross-modal thought chains \cite{ma2025cocova}; and Latent Implicit Visual Reasoning learns visual reasoning tokens without explicit intermediate supervision \cite{li2025livr}. FantasyVLN \cite{zuo2026fantasyvln} extends this line to VLN by unifying textual, visual, and multimodal CoT branches within a shared implicit reasoning framework. It mainly addresses the high token cost and low efficiency of explicit multimodal CoT in long-horizon VLN. By contrast, our method does not formulate navigation as unified CoT learning; it directly learns action-conditioned visual dynamics from trajectory future observations and internalizes them into a recurrent Pilot Token for anticipatory control.

\section{Method}
\subsection{Task Definition}
We consider instruction-following Vision-and-Language Navigation in continuous 3D environments as an instruction-conditioned partially observable Markov decision process (POMDP) \cite{kaelbling1998planning}.
Formally, let $\mathcal{M}=(\mathcal{S},\mathcal{A},\mathcal{O},\mathcal{T},\Omega)$, where $s_t\in\mathcal{S}$ denotes the (latent) environment state at time $t$.
Given a natural-language instruction $\mathbf{x}=(w_1,\ldots,w_L)$, the agent receives an egocentric monocular observation $\mathbf{o}_t\in\mathcal{O}\subset\mathbb{R}^{H\times W\times C}$ (with $C\!=\!3$ for RGB and optionally $C\!=\!4$ for RGB-D), and outputs a low-level action $a_t\in\mathcal{A}$.
The environment evolves according to
\begin{equation}
s_{t+1}\sim \mathcal{T}(\cdot\mid s_t,a_t),\qquad
\mathbf{o}_{t+1}\sim \Omega(\cdot\mid s_{t+1}),
\label{eq:pomdp}
\end{equation}
where $\mathcal{T}$ and $\Omega$ are the state transition and observation models, respectively.
This interaction repeats over time, producing a trajectory
\begin{equation}
\tau=\bigl(\mathbf{x},\,\mathbf{o}_{1:T},\,a_{1:T}\bigr),\qquad
\mathbf{o}_{1:t}\triangleq(\mathbf{o}_1,\ldots,\mathbf{o}_t),\quad
a_{1:t}\triangleq(a_1,\ldots,a_t),
\label{eq:traj_def}
\end{equation}
and terminates when $a_t=\texttt{STOP}$ or $t=T_{\max}$.
We adopt the standard discrete action set: 
\begin{equation}
\mathcal{A}=\{\texttt{FWD},\,\texttt{LEFT},\,\texttt{RIGHT},\,\texttt{STOP}\},
\label{eq:action_space}
\end{equation}
where each action corresponds to a small motion primitive in continuous space.
Our goal is to learn a policy $\pi_\theta(a_t\mid \mathbf{x},\mathbf{o}_{1:t})$ such that its rollouts reach the goal region $\mathcal{G} \subset\mathcal{S}$ and terminate by predicting $\texttt{STOP}$ at the target.

\begin{figure}[t!] 
    \centering
    \includegraphics[width=\linewidth]{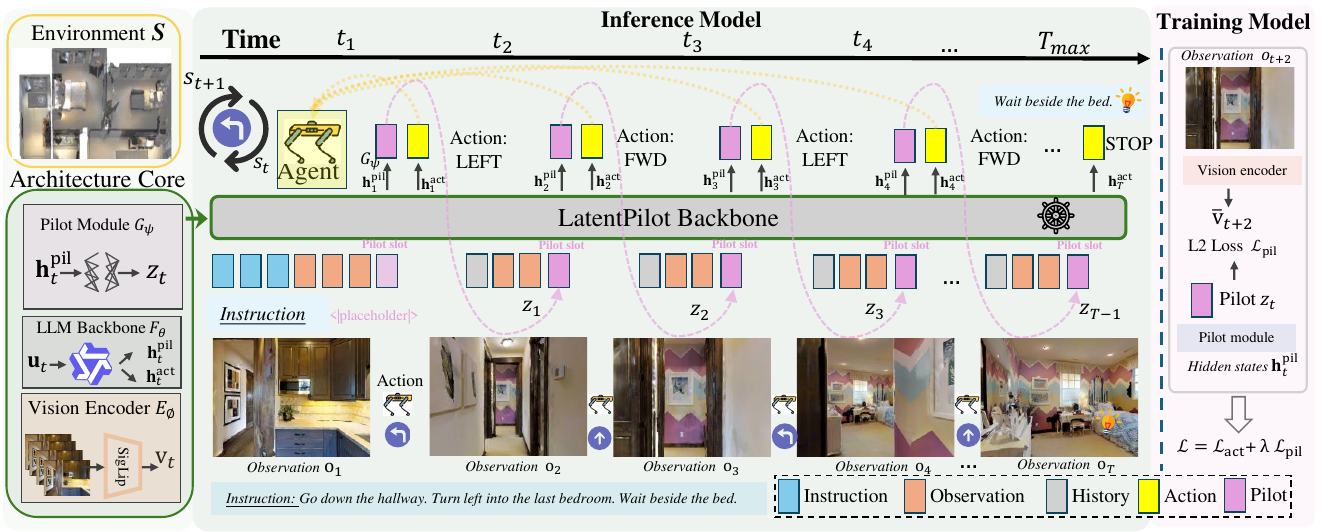}
    \vspace{-5mm}
\caption{Overview of LatentPilot. The vision encoder extracts visual tokens from the current observation, the LLM backbone predicts the action, and a lightweight Pilot module updates a propagated Pilot Token that is cached and reused as the Pilot slot input at the next step, enabling strictly causal dreaming-ahead.}
    \vspace{-2.5mm}
    \label{fig:pipeline}
\end{figure}
\subsection{LatentPilot}
\label{subsec:latentpilot}
As illustrated in Fig.~\ref{fig:pipeline}, LatentPilot is an end-to-end  navigator composed of three components: a vision encoder $E_{\phi}$, a lightweight Pilot module $G_{\psi}$, and a LLM decision backbone $F_{\theta}$.
The core idea is to maintain a step-propagated continuous latent state $\mathbf{z}_t\in\mathbb{R}^{d}$, termed the Pilot Token.
Pilot Token serves as a compact, persistent internal reasoning state throughout navigation, enabling the policy to carry forward anticipatory cues without generating verbose textual rationales.

\paragraph{Pilot initial.}
Given the egocentric observation $\mathbf{o}_t\in\mathbb{R}^{H\times W\times C}$ at step $t$, we adopt a SigLIP~\cite{zhai2023siglip}-initialized vision encoder to obtain a sequence of visual tokens
\begin{equation}
\mathbf{v}_t = E_{\phi}(\mathbf{o}_t)\in\mathbb{R}^{N_v\times d},
\label{eq:vision_tokens}
\end{equation}
where $N_v$ is the number of visual tokens and $d$ matches the hidden size of the backbone.
We initialize the decision backbone $F_{\theta}$ from the 7B LLaVA-Video model~\cite{zhang2024llavavideo}.
At each step, the backbone consumes a concatenated multimodal sequence $\mathbf{u}_t$ consisting of instruction tokens, current visual tokens, and the Pilot Token $\mathbf{z}_{t-1}$carried from the previous step:
\begin{equation}
\mathbf{u}_t \triangleq \Bigl[\ \mathrm{Tok}(\mathbf{x})\ ;\ \mathbf{v}_t\ ;\ \texttt{PILOT}\!\left(\mathbf{z}_{t-1}\right)\ \Bigr].
\label{eq:input_seq}
\end{equation}
Here $\texttt{PILOT}(\cdot)$ denotes placing $\mathbf{z}_{t-1}$ at a dedicated token positions, called Pilot slot. Pilot slot is implemented by a special placeholder token \texttt{<|placeholder|>} whose embedding provides $\mathbf{z}_0$ at the first step.
Unlike discrete language tokens, $\mathbf{z}_t$ lies in a continuous latent space and is carried across steps, making it a natural vehicle for long-horizon, trajectory-level internal computation.

\paragraph{Pilot propagation.}
The LLM backbone computes hidden states with causal attention,
\begin{equation}
\mathbf{H}_t = F_{\theta}(\mathbf{u}_t)\in\mathbb{R}^{N\times d}.
\label{eq:hidden}
\end{equation}
Let $\mathbf{h}^{\mathrm{act}}_t$ and $\mathbf{h}^{\mathrm{pil}}_t$ denote the action and Pilot-position hidden states from $\mathbf{H}_t$, respectively.
We obtain the low-level action distribution via a linear action head,
\begin{equation}
\pi_{\theta}(a_t \mid \mathbf{x},\mathbf{o}_t,\mathbf{z}_{t-1})
= \mathrm{Softmax}\!\left(\mathbf{W}_a\,\mathbf{h}^{\mathrm{act}}_t\right),
\quad a_t\in\mathcal{A}.
\label{eq:action_head}
\end{equation}
The Pilot Token is then updated by a lightweight projection-only Pilot module $G_{\psi}$:
\begin{equation}
\mathbf{z}_t = G_{\psi}\!\left(\mathbf{h}^{\mathrm{pil}}_t\right)\in\mathbb{R}^{d},
\label{eq:pilot_update}
\end{equation}
where $G_{\psi}$ is implemented as a simple linear layer, with only a few parameters that projects the Pilot-position hidden state back to the latent space.
Importantly, $G_{\psi}$ does not introduce any additional observations or external predictive components (e.g., world-model rollouts); it merely performs a compact latent projection to support cross-step propagation. Overall, one navigation step can be summarized as
\begin{equation}
(a_t,\mathbf{z}_t)=\mathcal{F}_{\theta,\psi}\bigl(\mathbf{x},\mathbf{o}_t,\mathbf{z}_{t-1}\bigr),
\label{eq:recurrent_form}
\end{equation}
where $\mathcal{F}_{\theta,\psi}$ denotes the overall step transition induced by the model: given the instruction, the current observation, and the previous Pilot Token, it outputs the next action and the updated Pilot Token.
The action $a_t$ is executed in the environment to yield the next observation $\mathbf{o}_{t+1}$, while $\mathbf{z}_t$ is fed into the next step through the Pilot slot.
This recurrent latent propagation links stepwise decisions into a coherent, trajectory-level internal reasoning process within a single backbone. So far we have described the architecture and information flow of LatentPilot.
In the next subsection, we introduce how training-time supervision derived from future observations shapes $\mathbf{z}_t$ into an anticipatory latent that captures action-conditioned visual dynamics, while keeping test-time inputs unchanged. $\bar{\mathbf{v}}_{t+1}$ and $\mathbf{z}_t$ is supervised toward $\bar{\mathbf{v}}_{t+2}$. 

\begin{figure}[t!] 
    \centering
    \includegraphics[width=0.95\linewidth]{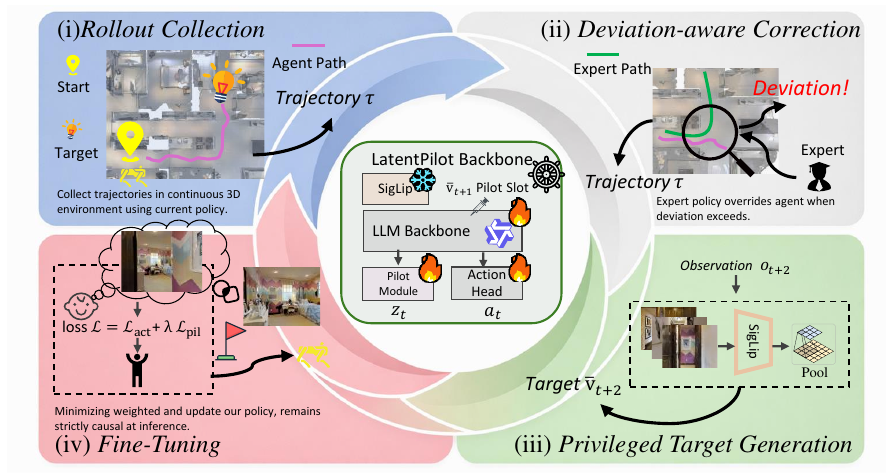}
    \vspace{-1mm}
\caption{PilotLoop: flywheel-style closed-loop training. Each round iterates rollout collection, expert takeover for deviation-aware correction, future-privileged target construction, and fine-tuning with joint action imitation and Pilot supervision.}
    \vspace{-5mm}
    \label{fig:pilotloop}
\end{figure}

\subsection{Training with Future-Privileged Supervision}
\label{subsec:future_privileged}
We train LatentPilot in a flywheel-style closed-loop procedure (Fig.~\ref{fig:pilotloop}), where data collection and model updates iterate in multiple rounds.
Each round cycles through four stages: (i) \emph{rollout collection} under the current policy, (ii) \emph{deviation-aware correction} where an expert takes over when the agent drifts too far from a reference trajectory, (iii) \emph{privileged target generation} from the collected rollouts, and (iv) \emph{fine-tuning} with a joint objective.
The key supervision signal comes from future observations recorded in these trajectories, used as training-only privileged information in the spirit of LUPI~\cite{vapnik2009new}.

\paragraph{Flywheel data collection with expert takeover.}
Let $\mathcal{D}=\{\tau^{(i)}\}_{i=1}^{N}$ denote the trajectory buffer collected in the current flywheel round, where
\begin{equation}
\tau = \bigl(\mathbf{x},\,\mathbf{o}_{1:T},\,a^{\mathrm{col}}_{1:T}\bigr).
\label{eq:traj_buffer}
\end{equation}
Here $a^{\mathrm{col}}_t$ is the collected action at step $t$. By default it is produced by the current model, while an expert policy may override it when the deviation from the reference becomes large~\cite{ross2011dagger}.
We then unroll LatentPilot along $\tau$ using the same recurrent interface as Eq.~\eqref{eq:recurrent_form} and optimize a standard action cross-entropy loss, here actions are decoded by the backbone's native LLM output projection.

\paragraph{Two-step future privilege.} A central design choice is to explicitly distinguish two future steps: $t\!+\!1$ as input, $t\!+\!2$ as target.
At training step $t$, we treat the next observation $\mathbf{o}_{t+1}$ as a privileged input to the Pilot slot, and the next-next observation $\mathbf{o}_{t+2}$ as a privileged target for supervising the predicted Pilot Token.
Concretely, we first compress an observation into a single latent vector by mean pooling the vision tokens:
\begin{equation}
\bar{\mathbf{v}}_{t+1} \triangleq \mathrm{Pool}\!\left(E_{\phi}(\mathbf{o}_{t+1})\right)\in\mathbb{R}^{d}.
\label{eq:pooled_vis}
\end{equation}

During training, we fill the Pilot slot with the one-step future latent $\bar{\mathbf{v}}_{t+1}$ as training-only privileged input. So multimodal input sequence $\mathbf{u}_t$ consisting of instruction tokens, current visual tokens, and one-step future:
\begin{equation}
\mathbf{u}^{\mathrm{tr}}_t \triangleq \Bigl[\ \mathrm{Tok}(\mathbf{x})\ ;\ \mathbf{v}_t\ ;\ \texttt{PILOT}\!\left(\bar{\mathbf{v}}_{t+1}\right)\ \Bigr],
\label{eq:train_input_priv}
\end{equation}
where $\mathbf{v}_t=E_{\phi}(\mathbf{o}_t)$ is the visual token sequence .
A forward pass yields hidden states $\mathbf{H}_t = F_{\theta}(\mathbf{u}^{\mathrm{tr}}_t)$ (Eq.~\eqref{eq:hidden}).
Let $\mathbf{h}^{\mathrm{act}}_t$ and $\mathbf{h}^{\mathrm{pil}}_t$ denote the action and Pilot-position hidden states from $\mathbf{H}_t$.
We supervise the action via:
\begin{equation}
p_{\theta}(\cdot)=\mathrm{Softmax}\!\left(\mathbf{W}_{\mathrm{LM}}\mathbf{h}^{\mathrm{act}}_t\right),
\qquad
\mathcal{L}_{\mathrm{act}}
\triangleq
-\sum_{t=1}^{T}\log p_{\theta}\!\left(a^{\mathrm{col}}_t \mid \mathbf{x},\mathbf{o}_t,\bar{\mathbf{v}}_{t+1}\right).
\label{eq:loss_act}
\end{equation}

To shape the Pilot Token into an anticipatory latent, we regress the predicted Pilot Token to match the two-step future latent. Concretely, the Pilot module projects the Pilot-position hidden state into the continuous latent space and the privileged target is defined from the two-step future observation:
\begin{equation}
\mathbf{z}_t \triangleq G_{\psi}\!\left(\mathbf{h}^{\mathrm{pil}}_t\right)\in\mathbb{R}^{d},
\qquad
\mathcal{L}_{\mathrm{pil}} \triangleq \sum_{t=1}^{T-2}\left\|\mathbf{z}_t-\bar{\mathbf{v}}_{t+2}\right\|_2^2.
\label{eq:loss_pil}
\end{equation}

\paragraph{Overall objective and train--test interface.}
We fine-tune the model in each flywheel round by minimizing
\begin{equation}
\min_{\theta,\phi,\psi}\;
\mathbb{E}_{\tau\sim\mathcal{D}}
\left[
\mathcal{L}_{\mathrm{act}}(\tau)
+\lambda\,\mathcal{L}_{\mathrm{pil}}(\tau)
\right],
\label{eq:loss_total}
\end{equation}
where $\lambda$ balances imitation learning and future-privileged supervision and we set 0.1 in practice.
Finally, note the train--test interface is simple and strictly causal at evaluation: during training, the Pilot slot is teacher-forced with 
We train LatentPilot on a mixed corpus of instruction-following trajectories from Matterport3D (MP3D) scenes, including R2R~\cite{anderson2018vln}, RxR~\cite{ku2020rxr}, and the EnvDrop-augmented R2R set~\cite{tan-etal-2019-learning}, and further include ScaleVLN trajectories synthesized on HM3D scenes~\cite{wang2023scalevln,ramakrishnan2021hm3d}.
We first bootstrap the policy with imitation learning using Habitat's shortest-path follower as the expert policy~\cite{savva2019habitat,krantz2020beyond}, and then perform flywheel-style closed-loop fine-tuning.


\begin{wrapfigure}{r}{0.5\linewidth}
    \centering
    \vspace{-20pt}
    \includegraphics[width=\linewidth]{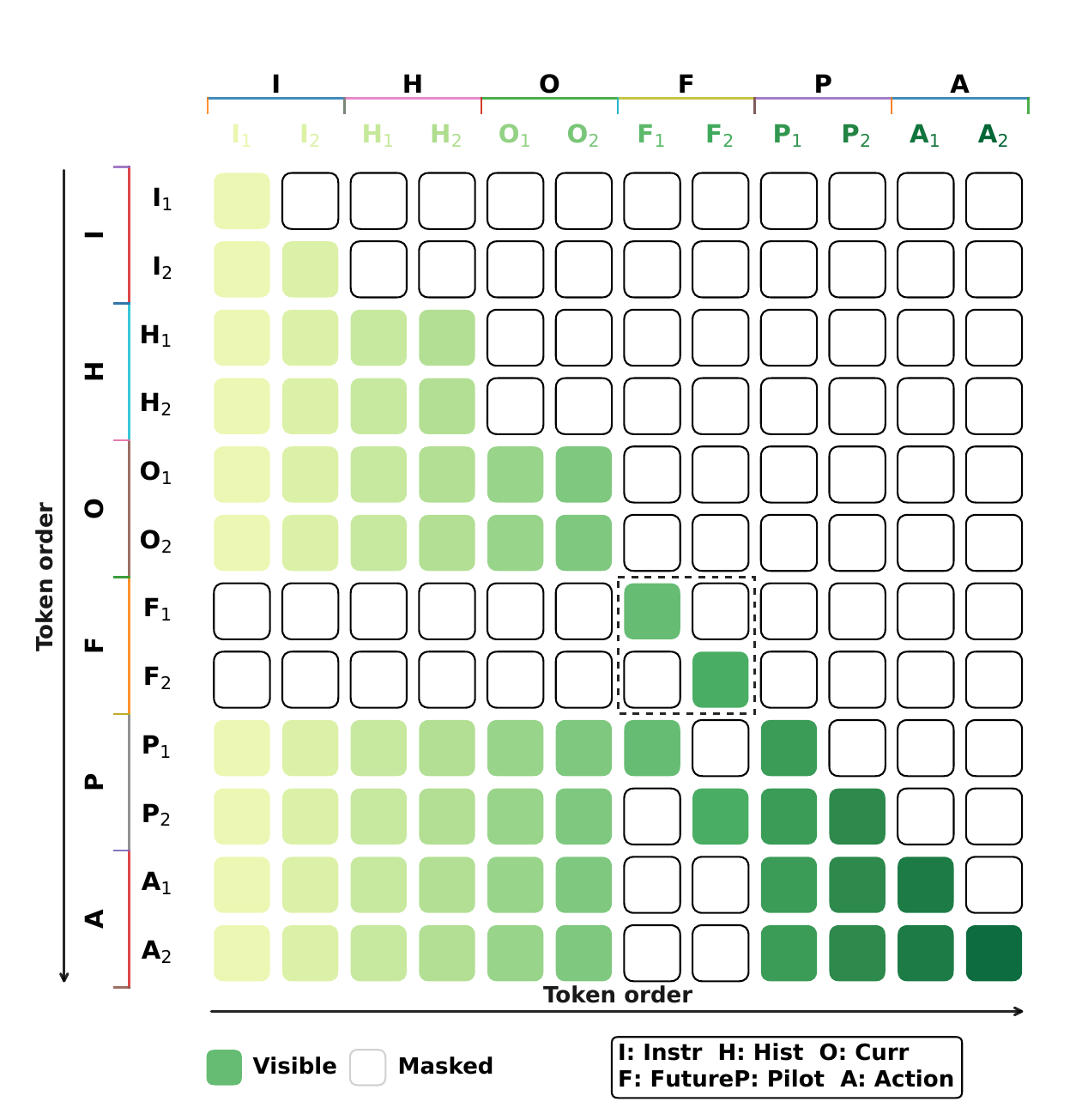}
    \vspace{-20pt}
    \caption{Visibility matrix of LatentPilot.}
    \vspace{-20pt}
    \label{fig:attension}
\end{wrapfigure}
\textbf{Inference with stored pilot.} During inference, the model performs a purely recurrent rollout a \texttt{PilotCache}, without accessing any future frames or computing additional Pilot representations. Instead of generating a separate Pilot signal, the model directly reuses the previously predicted Pilot Token through a read–write cache mechanism. 
Specifically, after step $t\!-\!1$ the model outputs $\mathbf{z}_{t-1}$, which is stored in the cache and used as the Pilot-slot input at step $t$ (i.e., $\texttt{PILOT}(\mathbf{z}_{t-1})$ in $\mathbf{u}_t$). The new output $\mathbf{z}_t$ then overwrites the cache for the next step. This read--write update repeats until \texttt{STOP}, enabling cross-step Pilot propagation using only the current observation and the cached latent.
Importantly, no separate Pilot computation is required at test time as the cached latent itself serves as the Pilot representation. As a result, cross-step Pilot propagation is achieved using only the current observation and a single cached latent.

\section{Experiments}
We conduct experiments to answer the following questions:
(1) Can we use these recorded future observations during training to learn how actions change future visual inputs, while keeping inference strictly causal? 
(2)What are the benefits of integrating future information and “imagination” directly into the decision-making LLM, compared to relying on external predictive models or auxiliary modules? We evaluate our method across representative robot embodiments, including \textbf{humanoid robots}, \textbf{wheeled platforms}, and \textbf{quadruped robots}, to demonstrate the broad applicability of LatentPilot. VLN-PE enables controlled experiments with simulated humanoid robots, and our real-world deployments further showcase LatentPilot on wheeled and quadruped robots.

\begin{table*}[!t]
\centering
\footnotesize
\setlength{\tabcolsep}{1.25pt}
\renewcommand{\arraystretch}{1.15}
    \vspace{-7pt}
\caption{Comparison with state-of-the-art methods on VLN-CE R2R and RxR \textbf{Val-Unseen} splits. Pano, Odo and D respectively represent panoramic view, odometry and depth, SRGB denotes monocular RGB.}
    \vspace{-10pt}
\resizebox{0.9\textwidth}{!}{%
\begin{tabular}{l|@{}c@{} @{}c@{} @{}c@{} @{}c@{}|cccc|ccc@{}@{}c@{}}
\toprule
\multirow{2}{*}{Method} &
\multicolumn{4}{c|}{Observation} &
\multicolumn{4}{c|}{R2R Val-Unseen} &
\multicolumn{4}{c}{RxR Val-Unseen} \\
\cmidrule(lr){2-5}\cmidrule(lr){6-9}\cmidrule(lr){10-13}
& Pano. & Odo. & D. & SRGB
& NE$\downarrow$ & OS$\uparrow$ & SR$\uparrow$ & SPL$\uparrow$
& NE$\downarrow$ & SR$\uparrow$ & SPL$\uparrow$ & nDTW$\uparrow$ \\
\midrule

CMA$ $~\cite{hong2022bridging}
& $\checkmark$ & $\checkmark$ & $\checkmark$ &
& 6.20 & 52.0 & 41.0 & 36.0
& 8.76 & 26.5 & 22.1 & 47.0 \\

VLN-BERT$ $~\cite{hong2022bridging}
& $\checkmark$ & $\checkmark$ & $\checkmark$ &
& 5.74 & 53.0 & 44.0 & 39.0
& 8.98 & 27.0 & 22.6 & 46.7 \\

Sim2Sim$ $~\cite{krantz2022sim2sim}
& $\checkmark$ & $\checkmark$ & $\checkmark$ &
& 6.07 & 52.0 & 43.0 & 36.0
& -- & -- & -- & -- \\

$\text{Ego}^2\text{-Map}$~\cite{hong2023ego2map}
& $\checkmark$ & $\checkmark$ & $\checkmark$ &
& 5.54 & 56.0 & 47.0 & 41.0
& -- & -- & -- & -- \\

DreamWalker~\cite{wang2023dreamwalker}
& $\checkmark$ & $\checkmark$ & $\checkmark$ &
& 5.53 & 59.0 & 49.0 & 44.0
& -- & -- & -- & -- \\

GridMM$ $~\cite{wang2023gridmm}
& $\checkmark$ & $\checkmark$ & $\checkmark$ &
& 5.11 & 61.0 & 49.0 & 41.0
& -- & -- & -- & -- \\

ETPNav$ $~\cite{an2023etpnav}
& $\checkmark$ & $\checkmark$ & $\checkmark$ &
& 4.71 & 65.0 & 57.0 & 49.0
& 5.64 & 54.7 & 44.8 & 61.9 \\

ScaleVLN$ $~\cite{wang2023scalevln}
& $\checkmark$ & $\checkmark$ & $\checkmark$ &
& 4.80 & -- & 55.0 & 51.0
& -- & -- & -- & -- \\

\midrule

InstructNav~\cite{long2024instructnav}
& $\checkmark$ & $\checkmark$ & $\checkmark$ & $\checkmark$
& 6.89 & -- & 31.0 & 24.0
& -- & -- & -- & -- \\

R2R-CMTP~\cite{chen2021cmtp}
& $\checkmark$ & $\checkmark$ & $\checkmark$ &
& 7.90 & 38.0 & 26.4 & 22.7
& -- & -- & -- & -- \\

LAW~\cite{raychaudhuri2021law}
&  & $\checkmark$ & $\checkmark$ & $\checkmark$
& 6.83 & 44.0 & 35.0 & 31.0
& 10.90 & 8.0 & 8.0 & 38.0 \\

CM2~\cite{georgakis2022cm2}
&  & $\checkmark$ & $\checkmark$ & $\checkmark$
& 7.02 & 41.5 & 34.3 & 27.6
& -- & -- & -- & -- \\

WS-MGMap~\cite{chen2022wsmgmap}
&  & $\checkmark$ & $\checkmark$ & $\checkmark$
& 6.28 & 47.6 & 38.9 & 34.3
& -- & -- & -- & -- \\

Sim2Real~\cite{wang2024sim2real3dff}
&  & $\checkmark$ & $\checkmark$ & $\checkmark$
& 5.95 & 55.8 & 44.9 & 30.4
& 8.79 & 25.5 & 18.1 & -- \\

Seq2Seq~\cite{krantz2020beyond}
&  &  & $\checkmark$ & $\checkmark$
& 7.77 & 37.0 & 25.0 & 22.0
& 12.10 & 13.9 & 11.9 & 30.8 \\

CMA~\cite{krantz2020beyond}
&  &  & $\checkmark$ & $\checkmark$
& 7.37 & 40.0 & 32.0 & 30.0
& -- & -- & -- & -- \\

\midrule

NaVid~\cite{zhang2024navid}
&  &  &  & $\checkmark$
& 5.47 & 49.1 & 37.4 & 35.9
& -- & -- & -- & -- \\

AO-Planner~\cite{chen2025aoplanner}
& $\checkmark$ &  & $\checkmark$ &
& 5.55 & 59.0 & 47.0 & 33.0
& 7.06 & 43.3 & 30.5 & 50.1 \\

COSMO~\cite{zhang2025cosmo}
& $\checkmark$ &  &  &
& -- & 56.0 & 47.0 & 40.0
& -- & -- & -- & -- \\

MapNav~\cite{zhang2025mapnav}
&  &  &  & $\checkmark$
& 4.93 & 53.0 & 39.7 & 37.2
& -- & -- & -- & -- \\

NaVid-4D~\cite{liu2025navid4d}
&  &  & $\checkmark$ & $\checkmark$
& 5.99 & 55.7 & 43.8 & 37.1
& -- & -- & -- & -- \\

NavMorph~\cite{yao2025navmorph}
&  &  & $\checkmark$ & $\checkmark$
& 5.75 & 56.9 & 47.9 & 33.2
& 8.85 & 30.8 & 22.8 & 44.2 \\

Uni-NaVid~\cite{zhang2025uninavid}
&  &  &  & $\checkmark$
& 5.58 & 53.3 & 47.0 & 42.7
& 6.24 & 48.7 & 40.9 & -- \\

NaVILA~\cite{cheng2025navila}
&  &  &  & $\checkmark$
& 5.22 & 62.5 & 54.0 & 49.0
& 6.77 & 49.3 & 44.0 & 58.8 \\

StreamVLN~\cite{wei2025streamvln}
&  &  &  & $\checkmark$
& 4.98 & 64.2 & 56.9 & 51.9
& 6.22 & 52.9 & 46.0 & 61.9 \\

JanusVLN~\cite{zeng2025janusvln}
&  &  &  & $\checkmark$
& 4.78 & 65.2 & 60.5 & 56.8
& 6.06 & 56.2 & 47.5 & 62.1 \\

\rowcolor{LightGray}
\textbf{LatentPilot (Ours)}
&  &  &  & $\checkmark$
& 4.41 & 66.3 & 62.0 & 58.0
& 5.19 & 58.2 & 49.9 & 67.5 \\

\bottomrule
\end{tabular}}
\label{tab:comp-vlnce}
    \vspace{-10pt}
\end{table*}

\begin{table}[!tbhp]
    \renewcommand\arraystretch{1.3}
    \centering
    \setlength{\tabcolsep}{2.9pt}
    \caption{Evaluation Metrics on VLN-PE benchmark with physical locomotion controller. +: model is first trained on Habitat and fine-tuned on VLN-PE. $\dagger$: model is trained with data augmentation.}
    \footnotesize
    \resizebox{0.92\linewidth}{!}{%
    \begin{tabular}{l|cccccc|cccccc}
        \toprule
        \multirow{2}{*}{Method} & \multicolumn{6}{c|}{R2R Validation Seen} & \multicolumn{6}{c}{R2R Validation Unseen}\\
        & NE$\downarrow$ & FR$\downarrow$ & StR$\downarrow$ & OS$\uparrow$ & SR$\uparrow$ & SPL$\uparrow$
        & NE$\downarrow$ & FR$\downarrow$ & StR$\downarrow$ & OS$\uparrow$ & SR$\uparrow$ & SPL$\uparrow$\\
                \midrule
         \multicolumn{13}{c}{\texttt{Train-free Map-based Exploration and Navigation}} \\
        \midrule
        VLMaps~\cite{huang23vlmaps} 
            & -- & -- & -- & -- & -- & --
            & 6.98 & 23.00 & 0.00 & 20.00 & 20.00 & 12.70 \\
        \midrule

        \multicolumn{13}{c}{\texttt{Train on VLN-PE}} \\
        \midrule
        Seq2Seq~\cite{krantz2020beyond} 
            & 7.73 & 22.19 & 3.04 & 30.55 & 19.60 & 15.67 
            & 7.91 & 19.67 & 3.71 & 27.62 & 15.89 & 12.58 \\
            Seq2Seq+~\cite{krantz2020beyond} 
            & 7.54 & 26.11 & 5.59 & 31.93 & 19.58 & 15.13 
            & 7.64 & 21.82 & 5.12 & 30.47 & 18.13 & 14.06 \\
        CMA  ~\cite{krantz2020beyond}     & 7.59 & 23.71 & 3.19 & 34.94 & 21.58 & 16.10 & 7.98 & 22.64 & 3.27 & 33.11 & 19.15 & 14.05 \\
        CMA+   ~\cite{krantz2020beyond}   & 7.14 & 23.56 & 3.50 & 36.17 & 25.84 & 21.75 & 7.26 & 21.75 & 3.27 & 31.40 & 22.12 & 18.65 \\
        RDP   ~\cite{wang2025vlnpe}     & 6.76 & 27.51 & 1.82 & 38.60 & 25.08 & 17.07 & 6.72 & 24.57 & 3.11 & 36.90 & 25.24 & 17.73 \\
        \midrule

        \multicolumn{13}{c}{\texttt{Zero-shot Transfer Evaluation from VLN-CE}} \\
        \midrule
        Seq2Seq~\cite{krantz2020beyond} $\dagger$ & 7.62 & 20.21 & 3.04 & 19.30 & 15.20 & 12.79 & 7.18 & 18.04 & 3.04 & 22.42 & 16.48 & 14.11 \\
        CMA ~\cite{krantz2020beyond}$\dagger$     & 7.37 & 20.06 & 3.95 & 18.54 & 16.11 & 14.64 & 7.09 & 17.07 & 3.79 & 20.86 & 16.93 & 15.24 \\
        NaVid   ~\cite{zhang2024navid}         & 6.20 & 11.25 & 0.46 & 24.32 & 21.58 & 17.45 & 5.94 & 8.61  & 0.45 & 27.32 & 22.42 & 18.58 \\
        DualVLN    ~\cite{wei2025dualvln}        & 4.13 & 17.78 & 1.82 & 62.31 & 58.97 & 47.78 & 4.66 & 12.32 & 2.23 & 55.90 & 51.60 & 42.49 \\
        \rowcolor{LightGray}
        \textbf{Ours} & \textbf{4.10} & \textbf{11.09} & \textbf{1.08} & \textbf{63.87} & \textbf{59.01} & \textbf{49.65} & 
        \textbf{4.33} &\textbf{ 10.65} &\textbf{ 0.97 }&\textbf{ 60.31} &\textbf{ 56.42} & \textbf{47.74 }\\

        \bottomrule
    \end{tabular}%
    }
    \label{tab:vlnpe}
    \vspace{-0.3cm}
\end{table}
\subsection{Environment and Metrics.}
We evaluate LatentPilot in both the standard continuous VLN setting and a physically realistic navigation setting.
For continuous navigation, we follow the VLN-CE  \cite{krantz2020beyond} and report results on the Val-Unseen splits of \textbf{R2R-CE} \cite{anderson2018vln} and \textbf{RxR-CE} \cite{ku2020rxr} and executed in the Habitat simulator \cite{savva2019habitat}.
Following prior work, we use the standard VLN metrics.
Navigation Error \textbf{(NE)} measures the final geodesic distance (in meters) from the agent's stopping location to the goal.
Success Rate \textbf{(SR)} is the fraction of episodes where the agent stops within $3$ meters of the goal.
Oracle Success Rate \textbf{(OSR)} considers the closest point along the trajectory as the stopping point.
Success weighted by Path Length \textbf{(SPL)} \cite{anderson2018vln} accounts for both success and path efficiency.
 Normalized Dynamic Time Warping \textbf{(nDTW) }\cite{ilharco2019dtw} measures trajectory fidelity to the reference path.

We additionally report results on \textbf{VLN-PE} \cite{wang2025vlnpe} follow DualVLN  ~\cite{wei2025dualvln}.
VLN-PE evaluation is executed on the physically realistic simulation stack built on NVIDIA Isaac Lab \cite{mittal2025isaaclab}, which models robot dynamics and locomotion execution imperfections. We report four primary VLN metrics: NE, SR, OS and SPL. To explicitly measure physical robustness, we further report Fall Rate \textbf{(FR)} measures the frequency of robot falls and Stuck Rate \textbf{(StR)} occurrences where the agent is unable to move.
 For humanoid evaluation in VLN-PE, we adopt the Unitree H1 embodiment, a full-size humanoid platform, about $1.8\,\mathrm{m}$ tall and equipped with depth camera \cite{unitree_h1} in simulator.
We perform on-policy rollout collection using 16$\times$RTX 4090 GPUs and training using 8$\times$NVIDIA A100 GPUs.

\subsection{Main result}
\textbf{Results on VLN-CE benchmark.}
Table~\ref{tab:comp-vlnce} summarizes results on the VLN-CE benchmarks. 
We compare against classic recurrent baselines (Seq2Seq/CMA) as well as stronger pipelines that rely on richer observations such as panoramic views, odometry, and depth (e.g., ETPNav and ScaleVLN)~\cite{krantz2020beyond,an2023etpnav,wang2023scalevln}. 
Despite using only monocular RGB, LatentPilot achieves the strongest overall performance, surpassing prior single-RGB VLM navigators such as NaVid~\cite{zhang2024navid} and remaining competitive with several sensor-heavy methods. 
LatentPilot compares favorably with existing approaches across standard VLN benchmarks while requiring a simpler observation setup, supporting the motivation that internalizing action-conditioned foresight into the decision backbone can reduce myopic “step-and-see” behavior.

\textbf{Results on VLN-PE benchmark.}
Table~\ref{tab:vlnpe} reports results on VLN-PE~\cite{wang2025vlnpe}, which evaluates continuous navigation under physically realistic humanoid locomotion with explicit robustness metrics (falls and deadlocks).
Across all baselines our method achieves the best overall performance, especially on the challenging Val-Unseen split.
LatentPilot not only more accurate at reaching goals but also more stable and safer under realistic execution noise.
\begin{table}[!t]
    \renewcommand\arraystretch{1.5}
    \centering
    \setlength{\tabcolsep}{6pt}
    \fontsize{8}{7}\selectfont
     \caption{Ablation study of supervision modality for the Pilot Token. NaN removes any extra Pilot supervision and optimizes only the action cross-entropy, serving as a baseline that tests whether the propagated latent becomes predictive by itself.}
     \vspace{-2mm}
         \resizebox{0.8\linewidth}{!}{%
    \begin{tabular}{l|cccc|cccc}
        \toprule
        \multirow{2}{*}{\textbf{Modality}} & \multicolumn{4}{c|}{R2R Validation Unseen} & \multicolumn{4}{c}{R2R Validation Seen}\\
          & SR$\uparrow$ & SPL$\uparrow$ & NE$\downarrow$ &  OS$\uparrow$ &  SR$\uparrow$ & SPL$\uparrow$ &NE$\downarrow$ & OS$\uparrow$\\
        \midrule
        NaN &51.7&	47.1&	5.3	&57.0 &53.1	&46.9&	4.6&	60.7 \\
        3D & 50.3	&45.6&	5.4&	55.8&51.5&	45.1&	4.9	&59.9 \\
        Text&53.2&	48.7	&5.2&	59.6&54.9&	48.6	&4.5	&63.9\\
        \midrule
        Vision(ours)&62.0&58.0&4.4&66.3&65.1&	59.6	&4.3&	71.6\\
        \bottomrule
    \end{tabular}
    }
    \label{tab:ab_study}
    \vspace{-1mm}
\end{table}
\subsection{Ablation Study}
\textbf{Supervision modality for the Pilot Token.}
We keep the LatentPilot architecture and inference interface unchanged and vary only the supervision signal used to shape the Pilot Token (Table~\ref{tab:ab_study}).
NaN removes any extra Pilot supervision and optimizes only the action cross-entropy, serving as a baseline that tests whether the propagated latent becomes predictive by itself.
3D uses geometry-aware features extracted by a 3D foundation model (VGGT~\cite{wang2025vggt}) as the privileged target, and Text uses textual descriptions of future views generated by a strong VLM (Qwen2.5-VL~\cite{bai2025qwen25vl}) and then tokenized as supervision.
We find that supervision choice matters substantially: proxy targets such as 3D embeddings or text provide limited improvements and can even degrade performance.

\begin{table}[!t]
    \renewcommand\arraystretch{1.35}
    \centering
    \setlength{\tabcolsep}{5pt}
    \fontsize{8}{7}\selectfont
\caption{Internalized future supervision (ours) vs.\ plug-in video world models on R2R Val-Seen. T/Act: ms per action; Mem: peak GPU memory (GB).}
    \vspace{-2mm}
    \resizebox{0.8\linewidth}{!}{%
    \begin{tabular}{l|cc|cccc}
        \toprule
        \multirow{2}{*}{\textbf{World Model}} &
        \multirow{2}{*}{T/Act (ms)$\downarrow$ } &
        \multirow{2}{*}{Mem$\downarrow$ (GB) } &
        \multicolumn{4}{c}{R2R Val-Seen}\\
        \cmidrule(lr){4-7}
        & & & SR$\uparrow$ & SPL$\uparrow$&
        NE$\downarrow$ & OS$\uparrow$ \\
        \midrule
        Wan2.1 1.3B \cite{wan2025wan} & 2040 & 41.5 & 56.6&	51.7&	5.1&	64.5 \\
        CogVideoX1.5 5B \cite{yang2024cogvideox} & 5460 & 32.4 &  59.3 & 53.9 & 4.9 & 66.8 \\
        \midrule
        \textbf{Ours} & 130 & 22.8 & \textbf{65.1} & \textbf{59.6} &\textbf{ 4.3 }& \textbf{71.6}\\
        \bottomrule
    \end{tabular}%
    }
    \label{tab:ab_study_worldmodel}
    \vspace{-1mm}
\end{table}
\textbf{Internalized imagination vs.\ plug-in world models.}
To further answer Question~(2), we compare LatentPilot with a representative class of \emph{external} attach a video world model~\cite{wan2025wan,yang2024cogvideox} to predict future views and then use these predictions to guide action selection (Table~\ref{tab:ab_study_worldmodel}).
Although such plug-in designs can provide explicit foresight, they also introduce an additional generative backbone at inference time, substantially increasing per-step latency and memory footprint.
In contrast, LatentPilot \emph{internalizes} the action-observation dynamics into the decision backbone via training-only future supervision, so inference remains a single forward pass with a lightweight latent propagation.
As a result, our approach achieves markedly lower time-per-action and peak memory, while also delivering stronger navigation performance.
Integrated imagination are not only improved success but also faster and more resource-efficient deployment.

\begin{wrapfigure}{r}{0.6\linewidth}
    \centering
    \vspace{-20pt}
    \includegraphics[width=\linewidth]{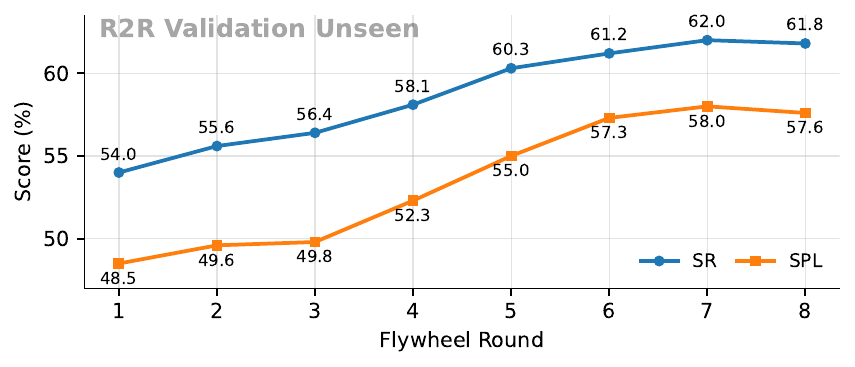}
    \vspace{-20pt}
    \caption{Performance over PilotLoop.}
    \vspace{-20pt}
    \label{fig:flywheel_curve}
\end{wrapfigure}

\textbf{Effect of flywheel iterations (PilotLoop).}
Figure~\ref{fig:flywheel_curve} plots performance on R2R Val-Unseen after each PilotLoop round. 
We observe a clear, steady improvement in both success (SR) and path efficiency (SPL) as the flywheel progresses, with the largest gains appearing in the early rounds and gradually saturating later. 
The expert-takeover mechanism stabilizes the loop by preventing severe drift when the policy deviates too far, echoing the dataset-aggregation intuition in DAgger-style training~\cite{ross2011dagger}. 

\begin{wrapfigure}{r}{0.55\linewidth}
    \centering
    \vspace{-20pt}
    \includegraphics[width=\linewidth]{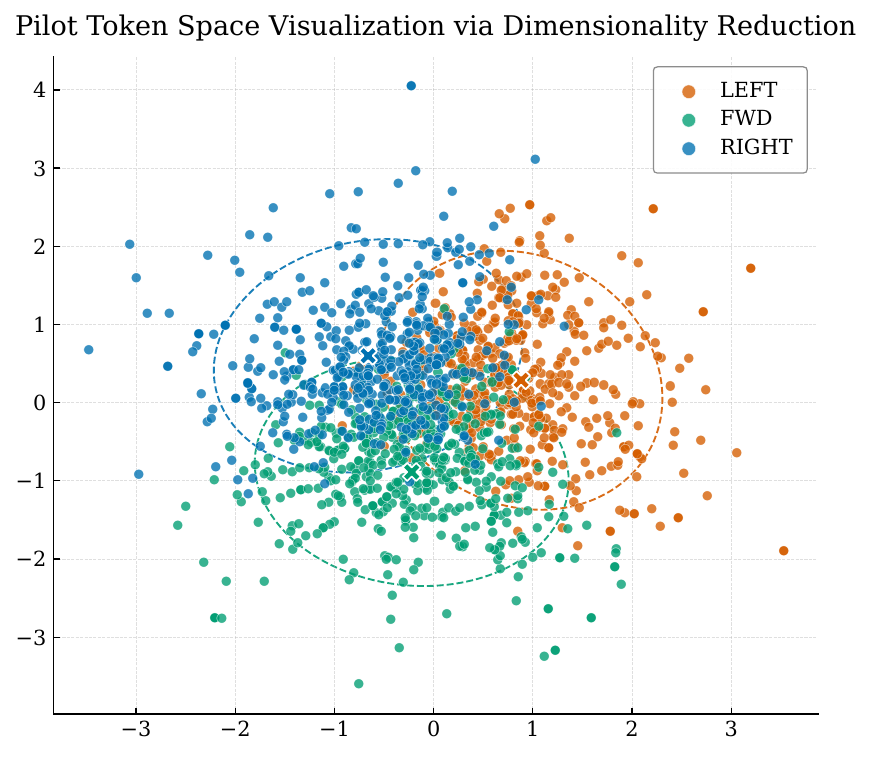}
    \vspace{-20pt}
    \caption{Latent collapse Analysis via PCA.}
    \vspace{-20pt}
    \label{fig:latent_pca}
\end{wrapfigure}
\textbf{Latent collapse analysis.}
A common failure mode of latent-token methods is latent collapse, where the latent vectors degenerate into near-constant, uninformative representations. 
To diagnose this, we collect the propagated Pilot Tokens $\mathbf{z}_t$ from rollouts and project them to 2D using principal component analysis (PCA)~\cite{jolliffe2002principal}. We then color each point by the executed action (\texttt{LEFT}, \texttt{FWD}, \texttt{RIGHT}); \texttt{STOP} is excluded due to its low frequency.
As shown in Fig.~\ref{fig:latent_pca}, Pilot Tokens form action-correlated and structured clusters rather than collapsing to a single mode, indicating that $\mathbf{z}_t$ preserves meaningful variation and remains strongly grounded in control-relevant information.
This supports that action grounding together with predictive supervision can maintain a non-degenerate latent space, enabling the Pilot Token to serve as a compact internal state instead of an empty placeholder.
\begin{figure}[!t] 
    \centering
    \includegraphics[width=\linewidth]{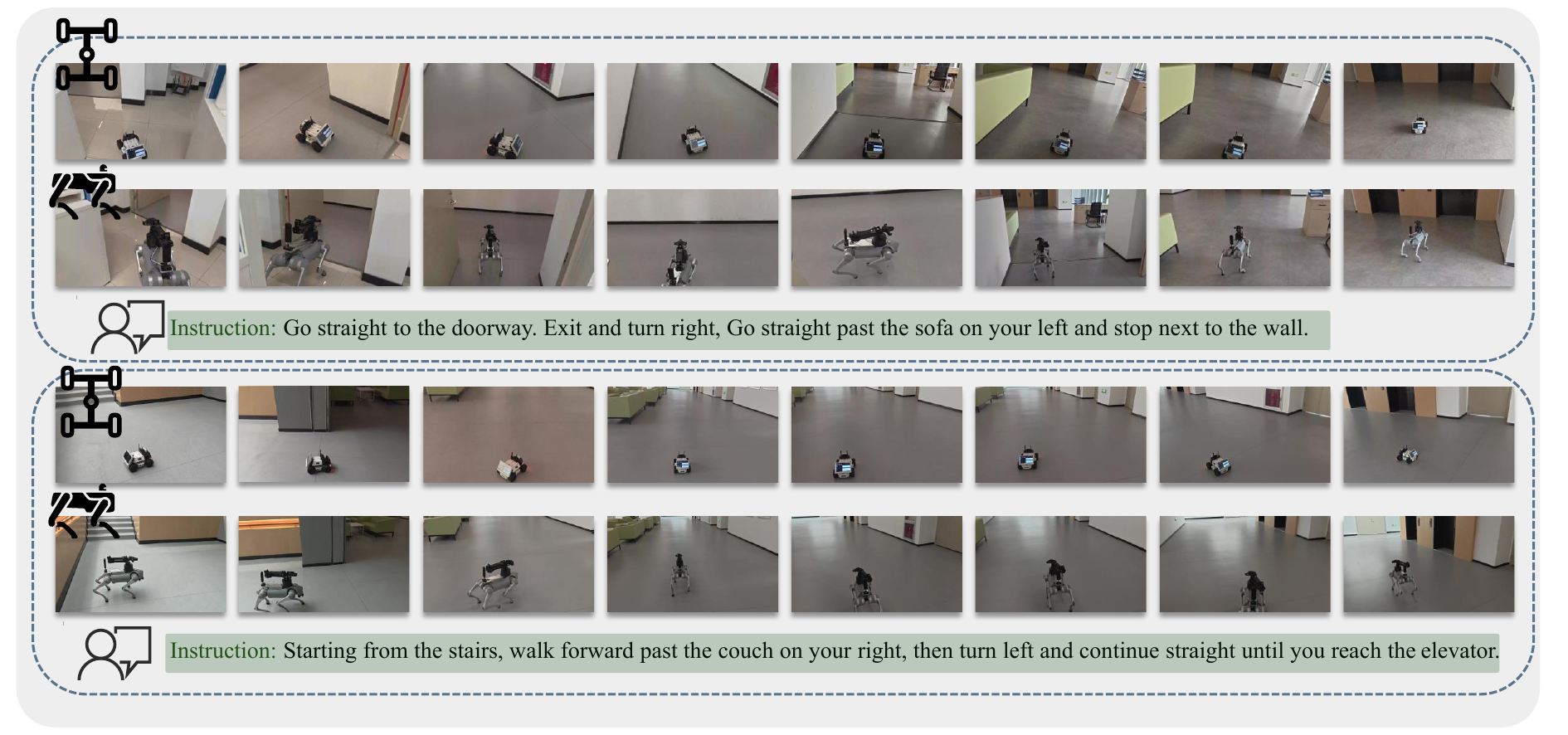}
    \vspace{-4mm}
    \caption{
Real-world Cross-Embodiment Experiments: Wheeled and Quadruped Robots.  LatentPilot successfully follows long, multi-step instructions on both Robots.}
    \vspace{-3mm}
    \label{fig:realrobot}
\end{figure}
\subsection{Real-world Experiments}
To further assess practical deployability beyond simulation, we deploy LatentPilot on two mainstream robot embodiments in indoor corridors: a wheeled AgileX LiMO Pro and a quadruped Unitree Go2.
LiMO Pro is equipped with LiDAR and a Intel Realsense D435 depth camera~\cite{agilex_limo_pro}, while Go2 is a bionic quadruped featuring a wide-angle camera.
Both robots run ROS~2 for I/O and control.
During execution, the robot streams egocentric observations to a remote workstation with RTX 4090 GPU over a local network; the server performs inference and sends back the predicted navigation action, which is then executed by the robot-specific low-level controller.
As shown in Fig.~\ref{fig:realrobot}, LatentPilot successfully follows long, multi-step instructions on both wheeled and legged platforms, suggesting that the learned Pilot Token provides a deployment-friendly form of internalized ``dreaming ahead'' without requiring any plug-in world-model rollout.
\section{Conclusion}
In this work, we presented LatentPilot, an end-to-end VLM-based navigator that internalizes lookahead for vision-and-language navigation.
Instead of relying on external imagination modules or world-model rollouts at inference, LatentPilot leverages a simple yet effective principle: although future observations are unavailable when an action is chosen, they are naturally recorded in offline trajectories and can serve as training-only privileged supervision.
By distilling these action-conditioned future visual consequences into a compact Pilot Token, our model learns anticipatory representations that support more stable sequential decision making while keeping inference strictly causal.

%
%
\clearpage
\bibliographystyle{splncs04}
\bibliography{main}
\clearpage
\includepdf[pages=-]{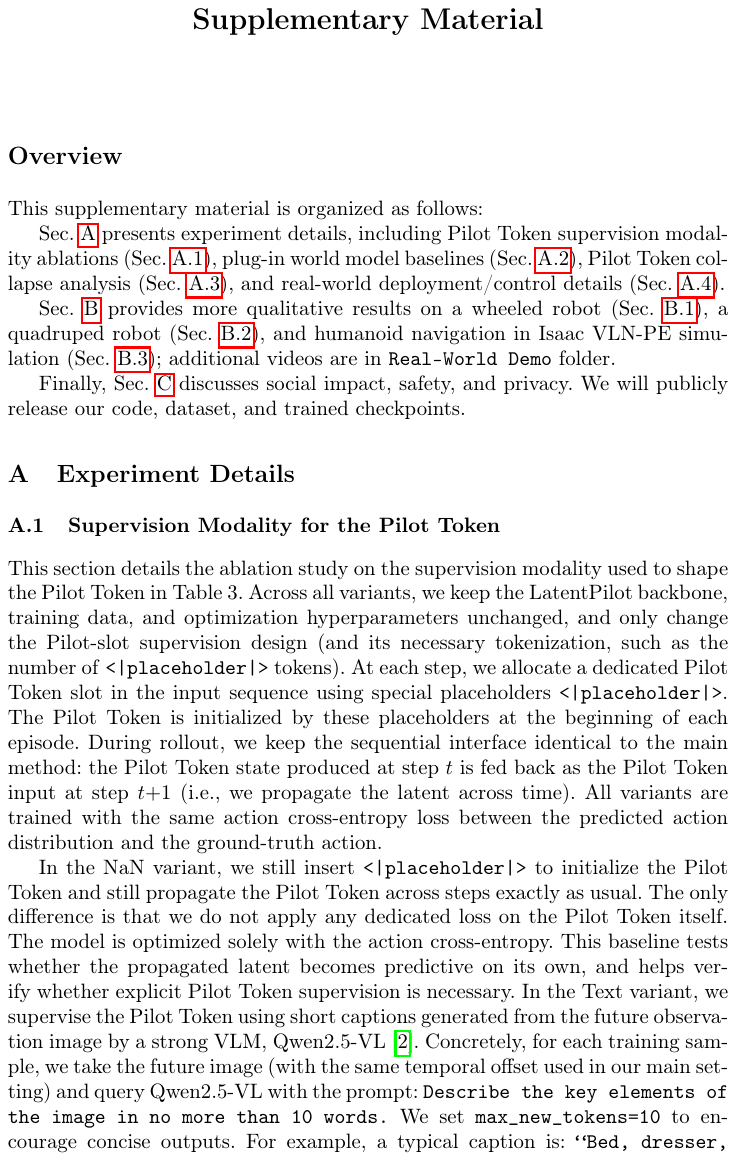}
\end{document}